\documentclass[a4paper, 10pt, conference]{ieeeconf}      % Use this line for a4 paper

\IEEEoverridecommandlockouts                              % This command is only needed if 
\usepackage{amsmath} % assumes amsmath package installed
\usepackage{amssymb}  % assumes amsmath package installed

\usepackage[tracking=true]{microtype}

\usepackage{graphicx}
\usepackage{moreverb,url}
\usepackage[colorlinks,bookmarksopen,bookmarksnumbered,citecolor=red,urlcolor=red]{hyperref}

\usepackage{dblfloatfix}    % To enable figures at the bottom of page
\usepackage{subfigure}
\usepackage{multirow}
\usepackage{booktabs}
\usepackage{xcolor}
\usepackage{stackengine}
\usepackage{tabularx}
\usepackage{colortbl}

\newcommand\xrowht[2][0]{\addstackgap[.5\dimexpr#2\relax]{\vphantom{#1}}}

\title{\LARGE \bf
Human-Robot Interface to Operate Robotic Systems \\ via Muscle Synergy-Based Kinodynamic Information Transfer*
}

\author{Janghyeon Kim$^{1}$, Dae Han Sim$^{1}$, Ho-Jin Jung$^{1}$, Ji-Hyeon Yoo$^{1}$, Changjae Lee$^{1}$, and Han Ul Yoon$^{2}$% <-this % stops a space
	\thanks{*This work was primarily supported by
		the National Research Foundation of Korea(NRF) grant funded
		by the Korea government (MSIT) (Grant No. 2021R1F1A1063339)
		and partially supported by 
		the MSIT National Program for ``Excellence in SW (Grant No. 2019-0-01219)'' supervised by
		the Institute of Information and communications Technology Planning and evaluation
		(IITP).
	}% <-this % stops a space
	\thanks{$^{1}$Janghyeon Kim, Dae Han Sim, Ho-Jin Jung, Ji-Hyeon Yoo, Changjae Lee are with the Department of Computer Science, Yonsei University~(Mirae), Wonju, Gangwon 26493, Korea
		\textls[40]{{\tt\small \{janghyeonk,dhsim,}} 
		\textls[40]{{\tt\small hojinj,jihyeonyoo,cjlee7128\}@yonsei.ac.kr}}   }%
	\thanks{$^{2}$Han Ul Yoon is with Faculty of the Division of Software, Yonsei University~(Mirae), Wonju, Gangwon 26493, Korea {\tt\small huyoon@yonsei.ac.kr}}%  \fontsize{15pt}{16.4pt}
}

\begin{document}

\maketitle
\thispagestyle{empty}
\pagestyle{empty}

%%%%%%%%%%%%%%%%%%%%%%%%%%%%%%%%%%%%%%%%%%%%%%%%%%%%%%%%%%%%%%%%%%%%%%%%%%%%%%%%
\begin{abstract}
%\color{blue}{
When a human performs a given specific task, it has been known that the central nervous system controls modularized muscle group, which is called muscle synergy. For human-robot interface design problem, therefore, the muscle synergy can be utilized to reduce the dimensionality of control signal as well as the complexity of classifying human posture and motion. In this paper, we propose an approach to design a human-robot interface which enables a human operator to transfer a kinodynamic control command to robotic systems. A key feature of the proposed approach is that the muscle synergy and corresponding activation curve are employed to calculate a force generated by a tool at the robot end effector. A test bed for experiments consisted of two armband type surface electromyography sensors, an RGB-d camera, and a Kinova Gen2 robotic manipulator to verify the proposed approach. The result showed that both force and position commands could be successfully transferred to the robotic manipulator via our muscle synergy-based kinodynamic interface.
%}
\end{abstract}

%%%%%%%%%%%%%%%%%%%%%%%%%%%%%%%%%%%%%%%%%%%%%%%%%%%%%%%%%%%%%%%%%%%%%%%%%%%%%%%%
\section{INTRODUCTION}
% 1st para
%In 1967, Bernstein found that the central nervous system sends modularized signals to control muscles when people perform a specific task\cite{bernstein1967co}. 
In 1967, Bernstein found that modularized muscle groups were coactivated by the central nervous system when people performed a specific task\cite{bernstein1967co}. 
For last two decades, studies have found the existence of the modularized muscle group for various tasks, e.g., balance maintenance\cite{ting2005limited}, swimming\cite{d2005shared}, reaching movements by hands\cite{berger2014effective}, grasping and picking an object\cite{furui2019myoelectric}, and so on. The modularized muscle group is called ``muscle synergy''\cite{d2006control,mckay2008functional,steele2013number}. The muscle synergy can be extracted from surface electromyography~(sEMG) by using non-negative matrix factorization\cite{lee2000algorithms}. 

% 2nd para
The muscle synergy can serve as a useful tool for human-robot interface design due to their characteristics, i.e., modularized and predetermined for a specific task. G\"unay et al. employed muscle synergy for hand posture classification; especially, a grasp classification for activities in daily living\cite{gunay2017muscle}. Furui et al. developed an impedance model-based biomimetic prosthesis control by utilizing the muscle synergy as motion primitives to determine combined motions\cite{furui2019myoelectric}. Those two research exemplify that the muscle synergy can contribute to reduce the dimensionality of control signal as well as the complexity of classification problem to identify various posture and motions, which are essential for human-robot interface design.  

% 3rd para
The interpretation of muscle synergy also gives us an information about the direction and magnitude of limb movement. Berger and d'Avella presented a method to translate a muscle synergy into a force in two dimensional Cartesian space\cite{berger2014effective}, and Camardella et al. proposed an approach to map a muscle synergy onto a force\cite{camardella2021towards}. Chen et al. showed that the activation curve of a muscle synergy can be used to calculate a force generated by a hand\cite{chen2020muscle}. All these existing findings substantiate that we can obtain a force-related information about a specific movement by analyzing the muscle synergy and corresponding activation curve. 
  
% 4th para
This paper proposes a human-robot interface design approach to operate robotic systems via kinodynamic information transfer. Specifically, both joint-and-skeleton and sEMG data are measured by RGB-d camera and armband type EMG sensor, respectively, while a pilot user is performing a specific task and then utilized to manipulate a robotic system with force and position commands.   
A key feature of our proposed design approach is that muscle synergies and corresponding activation curves are interpreted to determine the force command.  
%
%the joint-and-skeleton of a hand and upper limb sEMG data were measured by RGB-d camera and armband type EMG sensor, respectively, while a pilot user was performing a line drawing task with a digital pencil.
%
%We then extracted muscle synergy... to  force acting at the end of the digital pencil... which will be referred to as kinodynamic information transfer throughout the paper.
%
To our best knowledge, the human-robot interfaces via muscle synergy-based kinodynamic transfer have not been fully considered; therefore, findings from this study can contribute to the state of the arts in a field of the human-robot interface design.

% 5th para
{\color{black}{The rest of the paper is organized as follows: the proposed methodology to calculate muscle synergy-based kinodynamic information is introduced in Section 2. Specifically, our approach to obtain force and position command will be explained. In Section 3, the experimental test bed to verify the proposed approach is presented. The results are reported and significant outcomes are discussed in Section 4. Section 5 will be the conclusion of this paper.}}

\section{METHODS}

\subsection{Muscle Synergy: Definition}
\begin{figure}[t!]
	\centering
	\includegraphics[width=0.7\columnwidth]{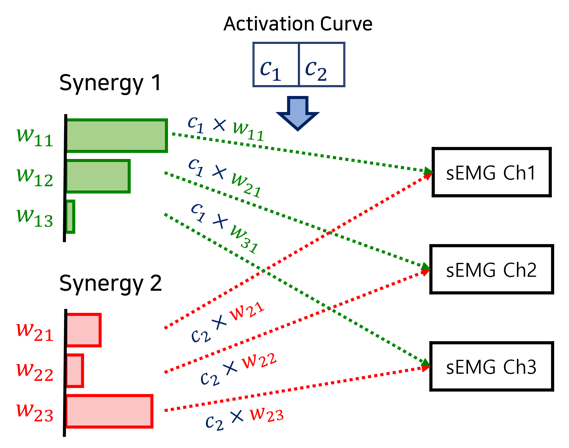}
	\caption{An illustrative example of the muscle synergy	in case of the numbers of channels and synergies are $d = 3$ and $n = 2$, respectively.}
	\label{Fig:fig1}
\end{figure}

Let $M \in \mathbb{R}^{d \times k}$ be $d$-channel sEMG signal for a time step $k$. Also, let a column vector $w_i=[w_{i1},w_{i2},\cdots,w_{id} ]^T \in \mathbb{R}^{d \times 1}$ be an $i^\mathrm{th}$ muscle synergy which is a modularized muscle group consisting of the contribution of each muscle. A row vector $c_i=[c_{i1},c_{i2},\cdots,c_{ik} ] \in \mathbb{R}^{1 \times k}$ denotes the corresponding activation curve of the muscle synergy $w_i$. Figure \ref{Fig:fig1} shows an illustrative example of the muscle synergy	in case of the number of channels and synergies are $d = 3$ and $n = 2$, respectively.
Consequently, $M$ can be expressed as follows:
%
%\[  M^{d \times k} = \sum_{i=1}^{n} w_i \times c_i \ = \ WC.   \]
%\[  M = \sum_{i=1}^{n} w_i \times c_i.  \]
\begin{equation}
	M = \sum_{i=1}^{n} w_i \times c_i.
\end{equation}
$w_i$ and $c_i$ can be found by applying the non-negative matrix factorization algorithm presented in~\cite{lee2000algorithms}.

%\subsection{Finding a Muscle Synergy of which Most Similarity to Human Generated Force}
%
%The template is used to format your paper and style the text. All margins, column widths, line spaces, and text fonts are prescribed; please do not alter them. You may note peculiarities. For example, the head margin in this template measures proportionately more than is customary. This measurement and others are deliberate, using specifications that anticipate your paper as one part of the entire proceedings, and not as an independent document. Please do not revise any of the current designations

%\subsection{Calculating a Force Generated by a Tool at a Robot End Effector} 

\subsection{Obtaining the Muscle Synergy-Based Kinodynamic Information} 
Procedure to obtain the muscle synergy-based kinodynamic information, i.e., the force and position commands for robotic systems, consists of the following three steps:
\begin{enumerate}
	\item Find a muscle synergy of which mostly correlated to human generated force.
	\item Calculate a force generated by a tool at the robot end effector.
	\item For position command, use the vector flow of the human operator hand's joint-and-skeleton data.  
\end{enumerate}
	
First, let $F_h \in \mathbb{R}^{1 \times k}$ be the magnitude of a force generated by a human operator for time step $k$ while performing a given specific task; for instance, the human operator is instructed to press down a solid surface with a tool in his hand. Recall that $w_i$ and $c_i$ represent a muscle synergy and a corresponding activation curve, respectively. Since $c_i$ is proportionally related to $F_h$, we can find the muscle synergy of which mostly correlated to the given specific task, denoted by $w_{F_h}$,
%
%\[ \textrm{arg} \max_i \{ F_h c_i^T \} \ \ \textrm{then} \ \  w_{F_h} \equiv w_i  \textrm{ and } c_{F_h} \equiv c_i  \]
%\[ \textrm{arg} \max_i \{ F_h c_i^T \} \ \ \textrm{then} \ \  w_{F_h} \equiv w_i, \]
\begin{equation}
	\textrm{arg} \max_i \{ F_h c_i^T \} \ \ \textrm{then} \ \  w_{F_h} \equiv w_i,
\end{equation}
and the corresponding activation curve can be determined straightforwardly as $c_{F_h} \equiv c_i$. 

Next, let $\widehat{F}$ represents the magnitude of force to be transferred to a robotic system, which the human operator wants to generate eventually by a tool at the robot end effector. $\widehat{F}$ can be calculated by
%
%\[ \widehat{F} = \alpha c_{F_h} \]
\begin{equation}
	\widehat{F} = \alpha c_{F_h}
\end{equation}
where $\alpha$ is a positive constant. We note that similar approaches can also be found in \cite{camardella2021towards,chen2020muscle}.

Lastly, the vector flow of the human operator hand's joint-and-skeleton data, which can be obtained by RGB-d camera, is directly used for position command. 

\begin{figure}[t!]
	\centering
	\includegraphics[width=0.95\columnwidth]{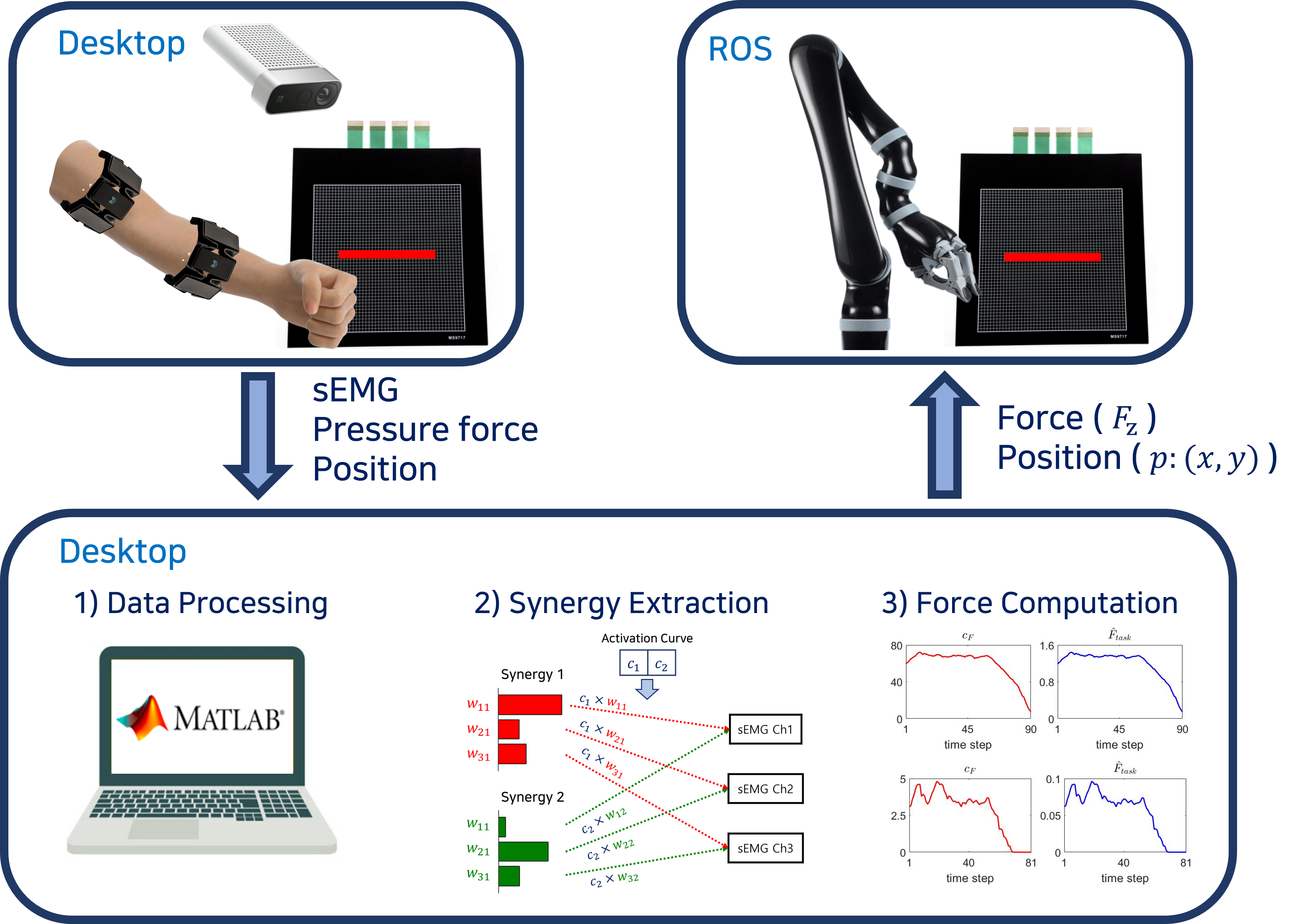}
	\caption{The system architecture of the proposed human-robot interface for kinodynamic information transfer.}
	\label{Fig:fig2}
\end{figure}

Figure \ref{Fig:fig2} presents the system architecture of the proposed human-robot interface.

\section{EXPERIMENTS}
\subsection{Pilot human operator and Given Task}
A pilot human operator was equipped with two armband type sEMG sensors on his upper arm and forearm, respectively. For each sEMG sensor, the channel deployment of the two sEMG sensors is summarized in Table \ref{Table:tbl1}. The pilot human operator was instructed to grip a digital pencil and draw a line from the left to the right five times by pressing a solid surface with two different force strengths: weak and strong; Hence, the pilot human operator drew a line for 10 trials.
\begin{table}[h!]
	\caption{The channel deployment of the two sEMG sensors}\label{Table:tbl1}
	\centering%
	{\footnotesize{
			\begin{tabular}{ c c c c }
				\toprule
				 & Ch \#1, \#2, \#3   &  Ch \#5, \#6, \#7      &  Ch \#4, \#8     \\ \midrule
				Upper Arm \xrowht{13pt} & Bicep Area  & Tricep Area  & \begin{tabular}[c]{@{}c@{}} Boundary \\ b/w channels \end{tabular}     \\  \midrule
				Forearm \xrowht{13pt} & Radius Side  & Ulna Side   &\begin{tabular}[c]{@{}c@{}} Boundary \\ b/w channels \end{tabular}     \\  \bottomrule
		\end{tabular}
	}}
\end{table}	

\subsection{Data Collection and Processing}
While the pilot human operator was performing the given task, sEMG data, the position of an index finger, and pressure force were measured by the two armband type sEMG sensors~(MyoArmband, Tahlmic Labs, Brooklyn, NY), an RGB-d camera~(Kinect  Azure, Microsoft, Redmond, WA), and a pressure mat~(MS9717, Kitronyx, Seoul, Republic of Korea), respectively. The $z$-axis of the RGB-d camera was aligned to the surface norm direction of the pressure mat. Under the pilot human operators demonstration, the robotic manipulator~(Kinova Gen2+KG3, Kinova, Quebec, Canada) drew a line on $xy$-plane.

The data processing was performed as follows. First, the sEMG data was moving average filtered, the pressure force was calculated by summing up all returned value from $16 \times 10$ cells, and the position of the index finger was recorded and the Kalman filter was applied. Next, after filtering processes, all data were up/down sampled to make them of the same length. Lastly, by following the presented approach in~\cite{mcdonald2021effect}, all data during 10 trials were concatenated. Consequently, we have
\begin{itemize}
	\item $M$: sEMG data $\mathbb{R}^{16 \times 939}$. The $1^\textrm{st}(9^\textrm{th})$ through $8^\textrm{th}(16^\textrm{th})$ row data corresponds to upper arm(forearm).
	\item $F_h$: the force generated by the pilot human operator $\mathbb{R}^{1 \times 939}$.
	\item $p$: the $(x,y)$ position of the index finger $\mathbb{R}^{2 \times 939}$.
\end{itemize}

\subsection{Muscle Synergy Extraction and Kinodynamic Information Calculation}
To calculate a force command $\widehat{F}$, muscle synergies were extracted from $M \in \mathbb{R}^{16 \times 939}$ under 90\% variation accounted for~(VAF)~\cite{lee2000algorithms,steele2013number}. From the extracted $w_i \in \mathbb{R}^{16 \times 1}$ and $c_i \in \mathbb{R}^{1 \times 939}$, we could find $w_{f_h}$ by (2). Afterward, $\widehat{F}$ was calculated by (3). $\alpha$ was set to 20. To distinguish two different pressure forces demonstrated by a pilot human operator, $\widehat{F}_\textrm{weak}$ and $\widehat{F}_\textrm{strong}$ will be used in Section IV.

As aforementioned, the position command $p$ could be directly obtained from RGB-d camera data. Finally, the  force generated by a tool at the robot end effector, denoted by $F_{r}$, was measured by the pressure mat. $F_h$ and $F_r$ were analyzed and compared each other.

%----------------------
\section{RESULTS}

\subsection{Result 1: Extracted Muscle Synergies and Force Command}
\begin{figure}[b!]
	\centering
	\includegraphics[width=0.95\columnwidth]{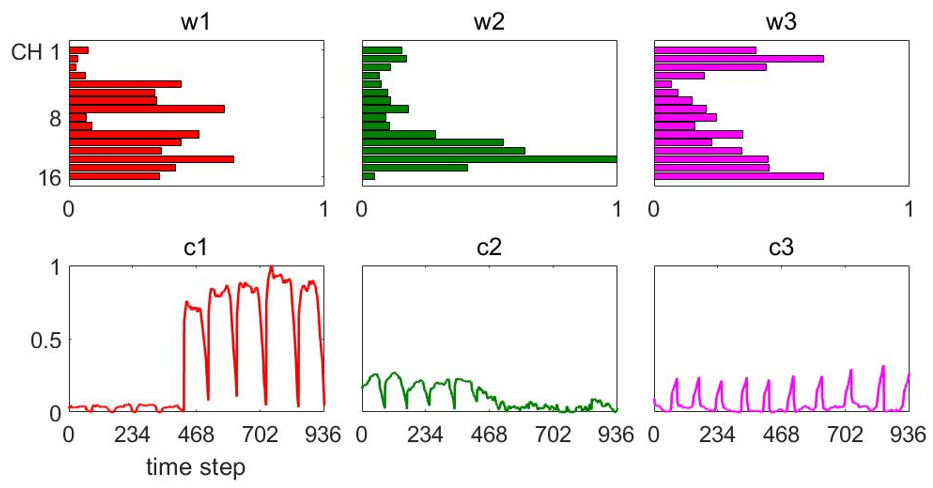}
	\caption{The extracted muscle synergies~(top row) and the corresponding activation curves~(bottom row) from the concatenated 10 trials sEMG data~($n=3$).}
	\label{Fig:fig3}
\end{figure}

Figure \ref{Fig:fig3} show the extracted muscle synergies~(top row) and the corresponding activation curves~(bottom row) from the concatenated 10 trials sEMG data under the number of synergies $n=3$. From the extracted muscle synergies, the size of bar graph represents the contribution of individual muscle for each synergy. The first half of the activation curves along time step shows the activation of the corresponding muscle synergy while a pilot user is drawing a line
with weak pressing force whereas the second half presents the activation under a strong pressing force. Both the muscle synergies and the corresponding activation curves are inter-channel normalized. 

\begin{figure}[t!]
	\centering
	\includegraphics[width=0.7\columnwidth,height=3.cm]{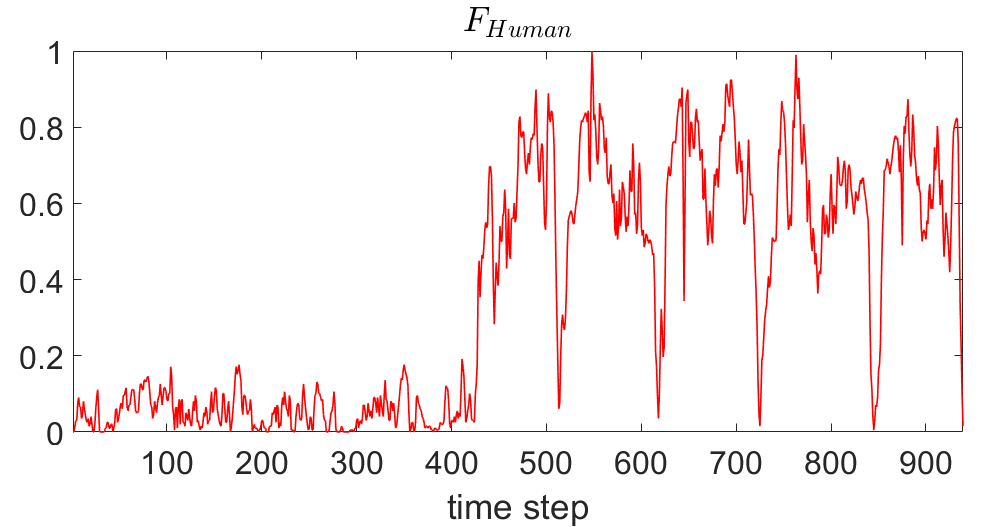}
	\caption{$F_h$ during a line drawing with a weak pressing force followed by a strong pressing force.}
	\label{Fig:fig4}
\end{figure}

Figure \ref{Fig:fig4} shows $F_h$ during a line drawing with a weak pressing force followed by a strong pressing force. By using (2), we could find that $w_{F_h} = w_1$ and accordingly $c_{F_h} = c_1$. We note that this result can be also observed by comparing $c_i$s in Fig. \ref{Fig:fig3} to $F_h$ in Fig. \ref{Fig:fig4}.   

\subsection{Result 2: Operating Kinova Gen2 Robotic Manipulator via Kinodynamic Information Transfer}
\begin{figure}[h!]
	\centering
	\includegraphics[width=0.85\columnwidth,height=3.75cm]{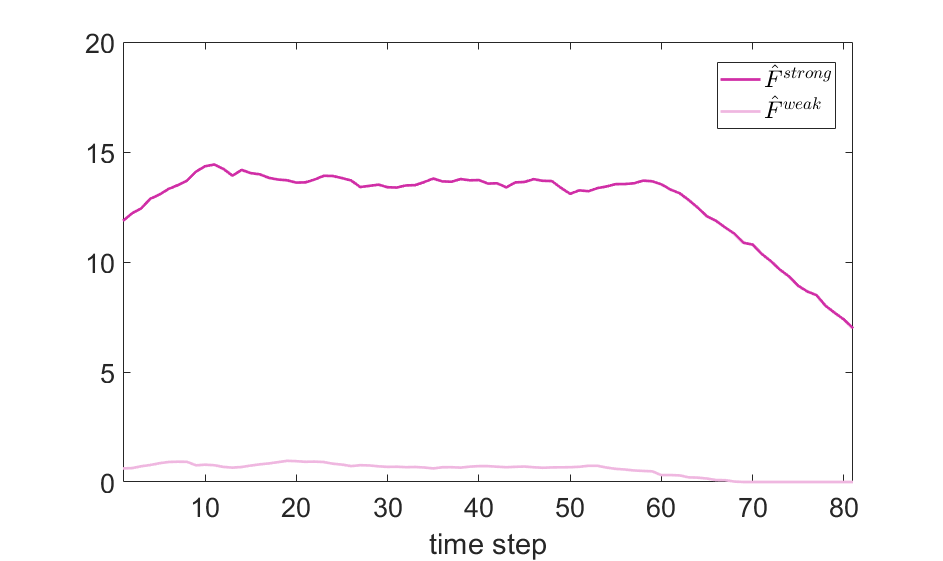}
	\caption{The force command $\widehat{F}_\textrm{weak}$ and $\widehat{F}_\textrm{strong}$.}
	\label{Fig:fig5}
\end{figure}

Figure \ref{Fig:fig5} depicts the force commands $\widehat{F}_\textrm{strong}$ and $\widehat{F}_\textrm{weak}$ for one trial which could be calculated using (3). We can see that $\widehat{F}_\textrm{strong}$ is ranging within $[0.0, 1.5]$ whereas $\widehat{F}_\textrm{weak}$ has values within $[0.0, 0.1]$. From Fig. \ref{Fig:fig5}, we could verify that $\widehat{F}_\textrm{strong}$ and $\widehat{F}_\textrm{weak}$ were successfully obtained for $F_h$ of different pressing force.

\begin{figure}[t!]
	\centering
	\includegraphics[width=0.85\columnwidth]{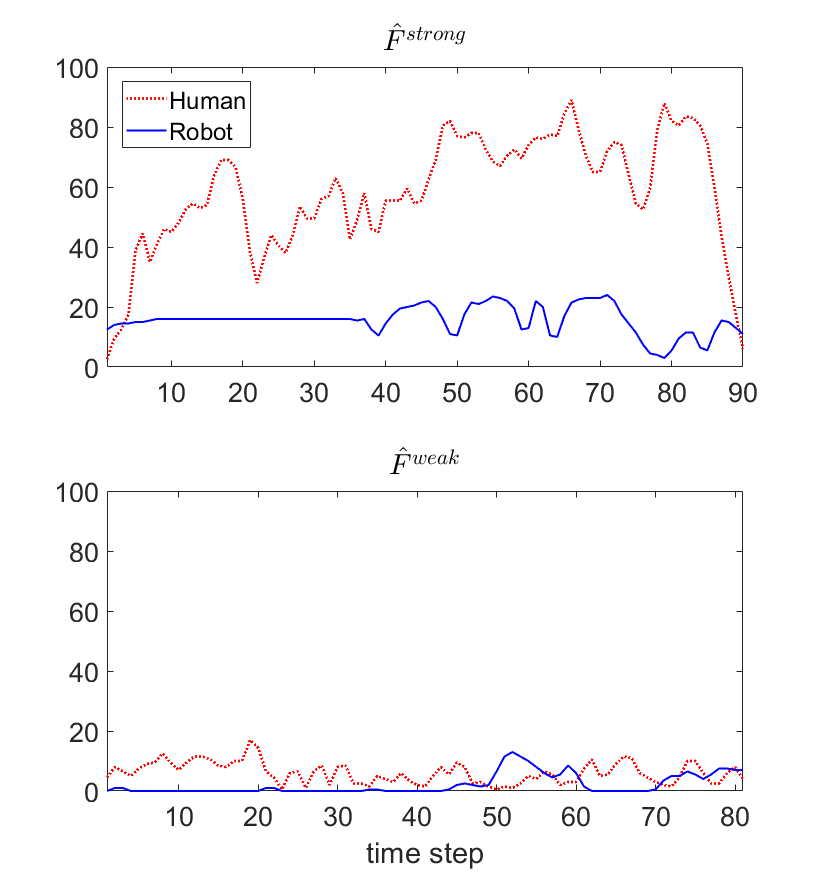}
	\caption{The comparison result of $F_h$ and $F_r$ for one trial when $\widehat{F}_\textrm{strong}$ and $\widehat{F}_\textrm{weak}$ were transferred to the robotic manipulator.}
	\label{Fig:fig6}
\end{figure}

Figure \ref{Fig:fig6} shows the comparison result of $F_h$ and $F_r$ for one trial when $\widehat{F}_\textrm{strong}$ and $\widehat{F}_\textrm{weak}$ were transferred to the robotic manipulator, respectively. As shown in the Fig. \ref{Fig:fig6}, the gain of the transferred force command could be adjusted by $\widehat{F}_\textrm{strong}$ and $\widehat{F}_\textrm{weak}$. It turned out that $\alpha=20$ in (3) was somewhat over-suppressed value for the force command transfer. We note that $\alpha$ was set to 20 to guarantee a safe force-operation range for our robotic manipulator.  

\begin{figure}[t!]
	\centering
	\includegraphics[width=0.85\columnwidth]{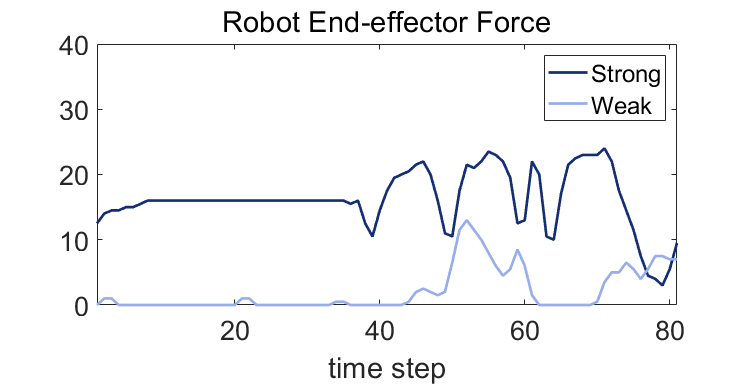}
	\caption{The comparison result of $F_r$ under $\widehat{F}_\textrm{strong}$ and $\widehat{F}_\textrm{weak}$.}
	\label{Fig:fig7}
\end{figure}

The comparison result of $F_r$ under $\widehat{F}_\textrm{strong}$ and $\widehat{F}_\textrm{weak}$ is presented in Fig. \ref{Fig:fig7}. From Fig. \ref{Fig:fig7}, we can also check that muscle synergy-based kinodynamic information was successfully transferred and enabled us to operate the robotic manipulator with generating different force by the tool at the end effector. 

\begin{figure}[t!]
	\centering
	\includegraphics[width=0.85\columnwidth]{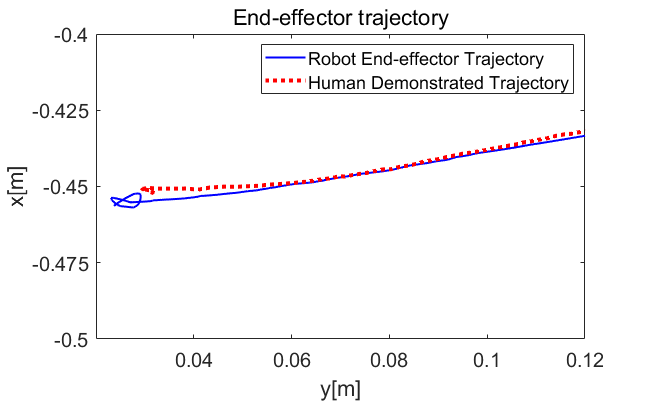}
	\caption{Robot manipulator end-effector and human demonstrated trajectories on $xy$-plane.}
	\label{Fig:fig8}
\end{figure}

Figure \ref{Fig:fig8} shows the robot manipulator end-effector and human demonstrated trajectories on $xy$-plane. From Fig.
\ref{Fig:fig8}, we can see that the human demonstration was well-transferred to the robotic manipulator end-effector, which substantiated kinematic information was successfully transferred via the proposed human-robot interface.

\addtolength{\textheight}{-0.85cm}   % This command serves to balance the column lengths

%%%%%%%%%%%%%%%%%%%%%%%%%%%%%%%%%%%%%%%%%%%%%%%%%%%%%%%%%%%%%%%%%%%%%%%%%%%%%%%%

%

\section{CONCLUSIONS}
Throughout this paper, we proposed an approach to design a human-robot interface to operate robotic systems via 
muscle synergy-based kinodynamic information transfer. First, in our interface design, a pilot human operator's both kinematic data and sEMG data were measured by armband type sEMG sensor and RGB-d camera, respectively. Next, muscle synergies were extracted from the measured sEMG data set and a specific muscle synergy, which was mostly related to the force generation by the human operator while performing a given task, was searched. Finally, the force and position commands (which were referred to as kinodynamic information) were transferred to a robotic manipulator. The experimental results showed that the pilot human operator was able to succefully perform a given line drawing task by applying various surface pressing forces under our proposed human-robot interface design approach.   

Future studies should be followed in the directions of relaxing the over-suppression of transfer gain parameter as well as challenging more complex task with the proposed interface design. Based on
findings in this study, we expect that the proposed design approach will be culminated to the
development of highly dexterous human-robot interface by which kinodynamic constrained tasks can be successfully performed.

%%%%%%%%%%%%%%%%%%%%%%%%%%%%%%%%%%%%%%%%%%%%%%%%%%%%%%%%%%%%%%%%%%%%%%%%%%%%%%%%

%%%%%%%%%%%%%%%%%%%%%%%%%%%%%%%%%%%%%%%%%%%%%%%%%%%%%%%%%%%%%%%%%%%%%%%%%%%%%%%%
%\section*{APPENDIX}
%\section*{ACKNOWLEDGMENT}
%
%The preferred spelling of the word acknowledgment. in America is without an e. after the g.. Avoid the stilted expression, One of us (R. B. G.) thanks . . ..  Instead, try R. B. G. thanks.. Put sponsor acknowledgments in the unnumbered footnote on the first page.

%%%%%%%%%%%%%%%%%%%%%%%%%%%%%%%%%%%%%%%%%%%%%%%%%%%%%%%%%%%%%%%%%%%%%%%%%%%%%%%%

%References are important to the reader; therefore, each citation must be complete and correct. If at all possible, references should be commonly available publications.

\bibliographystyle{ieeetr} % 다양한 스타일 많음: teeetr, plain, ...
\bibliography{kjhref.bib} % .bib 파일 제목과 같이 함

\end{document}